%% file: main.tex
\pdfoutput=1
\documentclass[letterpaper, 10 pt, journal, twoside]{IEEEtran}

\makeatletter
\newcommand{\dontusepackage}[2][]{%
  \@namedef{ver@#2.sty}{9999/12/31}%
  \@namedef{opt@#2.sty}{#1}}
\makeatother

\usepackage[utf8]{inputenc} 
\usepackage[T1]{fontenc}    
\usepackage{hyperref}       
\usepackage{url}            
\usepackage{booktabs}       
\usepackage{amsfonts}       
\usepackage{amsmath}
\usepackage{nicefrac}       
\usepackage{microtype}      
\usepackage{hyperref}
\usepackage[style=ieee]{biblatex}
\usepackage{multirow, rotating}
\usepackage{comment}

\usepackage{pifont}
\usepackage{graphicx}
\usepackage{xcolor}
\usepackage{longtable}
\usepackage{svg}
\usepackage{dsfont}
\usepackage{caption}
\usepackage{subcaption}
\usepackage[ruled,vlined]{algorithm2e}
\usepackage{enumerate}
\usepackage{soul}
\usepackage{adjustbox}

\hypersetup{
    colorlinks=true,
    linkcolor=black,
    filecolor=magenta,      
    urlcolor=blue,
    citecolor=black
}

\newcommand{\ie}{\emph{i.e.}~}
\newcommand{\eg}{\emph{e.g.}~}
\IEEEoverridecommandlockouts

\title{Embodied Visual Navigation with Automatic Curriculum Learning in Real Environments}

\date{February 2020}
\author{Steven D. Morad$^{1,2}$, Roberto Mecca$^{1}$, Rudra P.K. Poudel$^{1}$, Stephan Liwicki$^{1}$, and Roberto Cipolla$^{1,3}$%
\thanks{Manuscript received: September, 11, 2020; Revised October, 27, 2020; Accepted December, 11, 2020.}
\thanks{This paper was recommended for publication by Editor Tamim Asfour upon evaluation of the Associate Editor and Reviewers' comments. This work was supported by Toshiba Europe Limited.}%
\thanks{$^{1}$Steven D. Morad, Roberto Mecca, Rudra P.K. Poudel, Stephan Liwicki, and Roberto Cipolla are with Toshiba Europe Limited, Cambridge CB4 0GZ, UK {\tt\small \{firstname.lastname\}@crl.toshiba.co.uk}}%
\thanks{$^{2}$Steven Morad is with Computer Laboratory, University of Cambridge, 15 JJ Thomson Ave, Cambridge CB3 0FD, UK {\tt \small sm2558@cam.ac.uk}}%
\thanks{$^{3}$Roberto Cipolla is with the Department of Engineering, University of Cambridge, Trumpington St, Cambridge CB2 1PZ, UK {\tt\small rc10001@cam.ac.uk}}%
\thanks{Digital Object Identifier (DOI): see top of this page.}
}
\begin{document}
\maketitle
\setcounter{footnote}{3}


\begin{abstract}
    We present NavACL, a method of automatic curriculum learning tailored to the navigation task. NavACL is simple to train and efficiently selects relevant tasks using geometric features. In our experiments, deep reinforcement learning agents trained using NavACL significantly outperform state-of-the-art agents trained with uniform sampling -- the current standard. Furthermore, our agents can navigate through unknown cluttered indoor environments to semantically-specified targets using only RGB images. Obstacle-avoiding policies and frozen feature networks support transfer to unseen real-world environments, without any modification or retraining requirements. We evaluate our policies in simulation, and in the real world on a ground robot and a quadrotor drone. Videos of real-world results are available in the supplementary material.\footnote{Also available at \url{https://www.youtube.com/playlist?list=PLkG_dDkoI9pjPdOGyTec-sSu20pB7iayC}}
\end{abstract}
\input{sections/1_introduction}
\input{sections/2_approach}
\input{sections/3_model}
\input{sections/4_experiments}

\input{sections/5_conclusions}
\AtNextBibliography{\small}
\printbibliography

\end{document}

%% file: sections/1_introduction.tex
\section{Introduction}
\label{sec:introduction}




    
\IEEEPARstart{N}{avigation} forms a core challenge in embodied artificial intelligence (embodied AI) \cite{kadian2019we,8954627}. Before the embodied AI renaissance, approaches such as active vision \cite{aloimonos1988active} and active visual simultaneous localization and mapping (active VSLAM) \cite{leung2008active} were popular methods for building autonomous agents. They combined classical motion planning \cite{kavraki1996probabilistic,lavalle1998rapidly} with non-learned exploration policies such as frontier expansion \cite{yamauchi1997frontier} to direct the agent. Active VSLAM and active vision work well in ideal circumstances, but are brittle and lack generalization ability in real-world situations.
\begin{figure}
    \centering
    \begin{subfigure}{0.9\linewidth}
    \adjustbox{trim=0cm 0.3cm 0cm 0cm}{%
    \includegraphics[width=\linewidth]{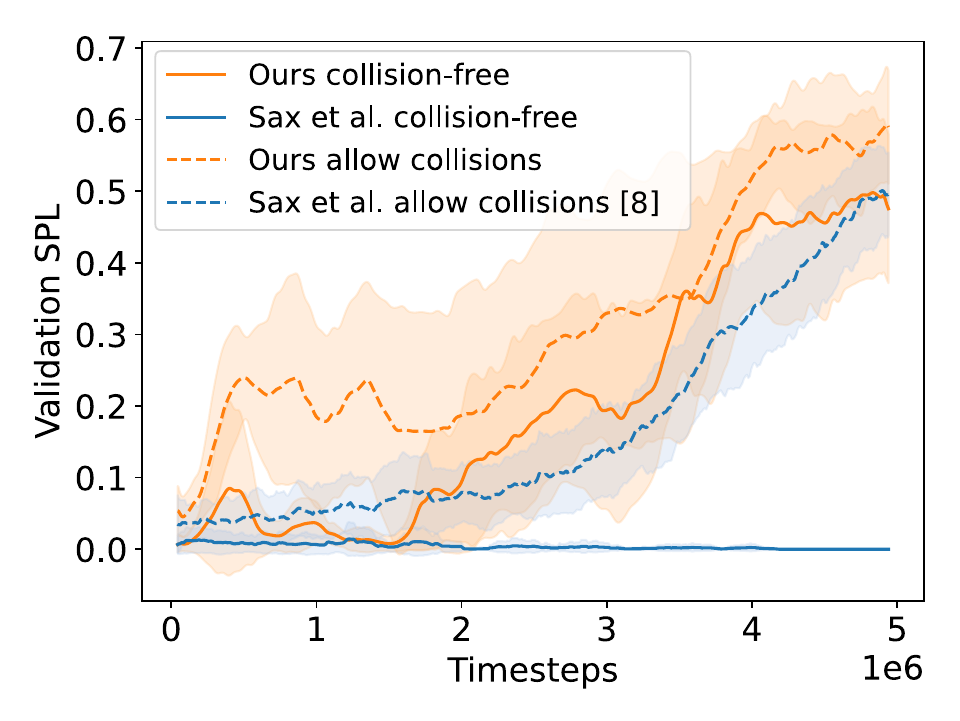}
    }
        \caption{Performance Comparison}
        \label{fig:stanford_plot}
    \end{subfigure}
    \begin{subfigure}{0.46\linewidth}%
\includegraphics[width=0.49\linewidth]{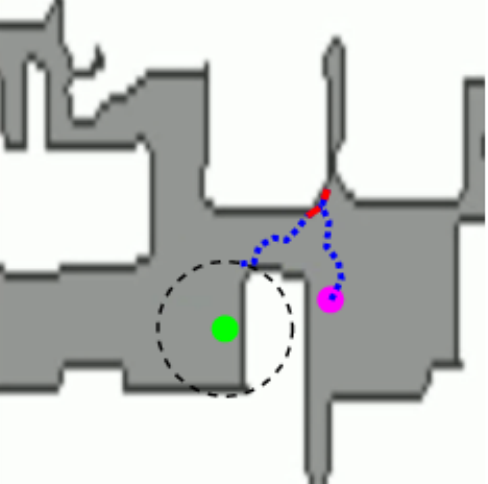}\hfill%
\includegraphics[width=0.49\linewidth]{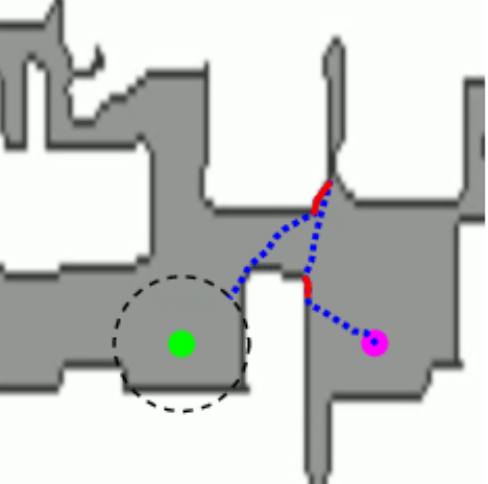}
\caption{Sax \emph{et al.} \cite{midLevelReps2018}}
        \label{fig:collisions_other}
    \end{subfigure}\hfill
    \begin{subfigure}{0.46\linewidth}%
\includegraphics[width=0.49\linewidth]{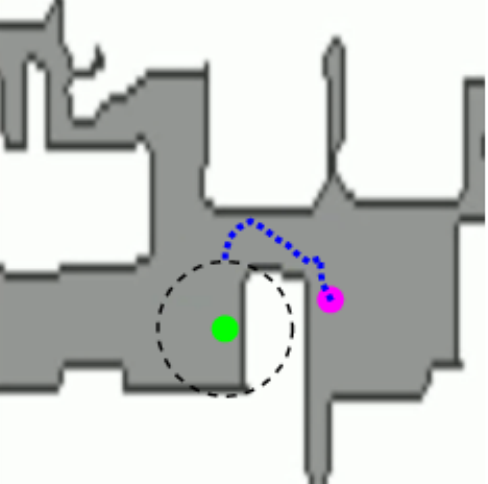}\hfill%
\includegraphics[width=0.49\linewidth]{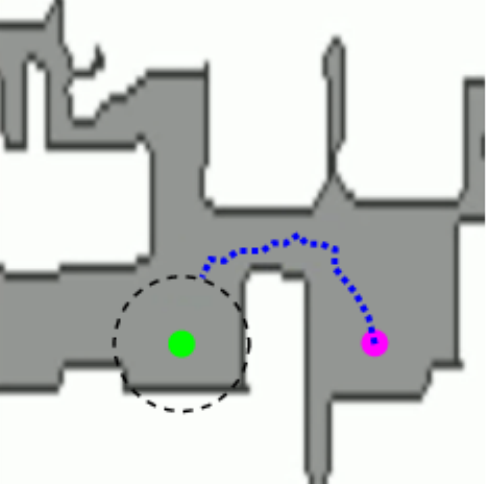}
\caption{Our Policy with NavACL}
        \label{fig:collisions}
    \end{subfigure}
    \caption{(\ref{fig:stanford_plot}) The 2019 CVPR Habitat RGB Navigation challenge winner \cite{midLevelReps2018} does not learn an effective policy when the collision-free property is enforced (episode termination and negative reward upon collision with an obstacle). Plots represent the mean and standard deviation of a single train/test scene over five trials. Collision-free agents are important for Sim2Real transfer but are much harder to train (SPL defined in eq. \ref{eqn:spl} -- larger is better). (\ref{fig:collisions_other}) Top-down view of fully-trained navigation agents exploiting collision mechanics to slide along walls. (\ref{fig:collisions}) Our agents with collision avoidance and automatic curriculum learning. The agent starts at the pink point and moves towards the green goal, producing a blue path. Red dots indicate collisions.}
    \label{fig:stanford_spl}
\end{figure}

Deep reinforcement learning (DRL) gained traction in the landmark paper \cite{mnih-atari-2013}, where DRL agents outperformed humans -- all be it in relatively simple arcade games. Since then, the scope of DRL has expanded to real-world applications. In robotics and navigation, DRL shows promise as an alternative to classical models due to its surprising robustness and ability to generalize to real-world uncertainties. \cite{mirowski2016learning} trained visual navigation agents in a video-game maze, showing that DRL agents can memorize the layout of a maze from vision alone. Since then, there has been an explosion of DRL-based visual navigation, fuelled by the abundance of indoor photo-realistic simulators and datasets \cite{xiazamirhe2018gibsonenv,straub2019replica,chang2018matterport3d,mo2018adobeindoornav,savva2017minos}. Evidence suggests DRL outperforms traditional methods in such cluttered, realistic indoor environments \cite{savva2019habitat}.

DRL visual navigation tasks can be broken down into point \cite{midLevelReps2018,chaplot2020learning,savva2019habitat}, object \cite{midLevelReps2018,mirowski2016learning,zhang2017deep}, or area \cite{mezghani2020learning,wu2019bayesian} navigation based on target description \cite{anderson2018evaluation}. Object navigation goals can be represented with images \cite{zhu2017target,mezghani2020learning} or semantics \cite{hill2020human,midLevelReps2018,mousavian2019visual}. Image-based representations match a reference image of the target object to the current agent observation, requiring a new image for each specific object instance. Semantic representations can be specified as a list of instructions \cite{wang2019reinforced}, or as a target label  \cite{mousavian2019visual}. Language instructions (\eg left at staircase) can preclude scene generalization. Semantic labels show great promise for real-world applications -- they are how humans describe targets to each other, and are not bound to a specific scene or object instance. Semantic label-based agents can be deployed to novel environments and duties; suitable for post-disaster recovery robots or embodied assistance technology with a wide range of task scenarios.

\subsection{Contributions}
We focus on indoor object-driven navigation to semantically-specified targets. We are interested in generalization to new environments and targets, including simulator-to-real-world (Sim2Real) generalization, without any retraining.

Utilizing semantic networks pretrained on large segmentation datasets like COCO \cite{lin2014microsoft}, we show agents with test-time generalization to object classes not seen in training simulations. To the best of our knowledge, we are the first to demonstrate this test-time generalization ability in simulation, and further extend it to the real world. We call this test-time generalization \emph{target-agnostic semantic navigation}.

While DRL visual navigation has proven itself in simulation, it rarely transfers to the real world. One challenge is overfitting to simulator-rendered images \cite{gordon2019splitnet}. Using frozen feature encoders trained on real images, \cite{midLevelReps2018} shows compelling generalization ability across multiple simulators. In our work, we demonstrate frozen feature networks and collision avoidance help bridge the Sim2Real gap by showcasing our policies on real robots in real environments. Another issue present in almost every navigation simulator is collision modeling exploitation \cite{kadian2019we}. Agents drive into a wall at an angle and slide along it, covering the perimeter of a simulated building (Fig. \ref{fig:collisions_other}). \cite{sadeghi2016cad2rl} demonstrates collision-avoidance policies trained entirely in simulation can transfer to the real world, but stops short of investigating longer-term navigation policies. DRL for long-term planning with obstacle avoidance has yet to be perfected.

Enforcing collision-free paths for agents results in increased reward sparsity, making training more difficult with state-of-the-art navigation tools (Fig. \ref{fig:stanford_plot}). We mitigate this elevated sparsity using automatic curriculum learning. The essence of curriculum learning is selecting and ordering training data in a way that produces desirable characteristics in the learner, such as generality, accuracy, and sample efficiency. Automatic curriculum learning (ACL) is the process of generating this curriculum without a human in the loop. Curriculum for neural networks was proposed by \cite{bengio2009curriculum}, and \cite{portelas2020automatic} affords a thorough overview of ACL applied to DRL. For navigation, tasks can be represented using low-dimensional Cartesian start and goal states. Some ACL methods that produce tasks of this form are asymmetric self-play \cite{sukhbaatar2017intrinsic} and GoalGAN \cite{florensa2018automatic}. Asymmetric self-play requires collecting distinct episodes for two separate policies, which is computationally expensive using 3D simulators. GoalGAN trades performance for generality. It can generate tasks for arbitrary problems, but uses a generative adversarial network (GAN) which is notoriously unstable and difficult to train. Instead, we trade generality for efficiency and propose a simple classification-based ACL method for navigation, termed NavACL.

In summary, we cast the visual navigation problem setup as follows:
\begin{enumerate}[i]
\vspace{-0.1cm}
    \item the agent's observations consist of RGB images from an agent-mounted camera and the semantic label of the target (\eg ``football'', ``vase''),
    \item the agent's actions consist of discrete, position-based motion primitives (\ie move forward, turn left or right), without explicit loop closure outside of said primitives,
    \item upon reaching the target, collision with the environment, or exceeding a preset time limit, the episode ends
\end{enumerate}
and contribute:
\begin{enumerate}[i]
\vspace{-0.1cm}
    \item a simple and efficient method to automatically generate curriculum for navigation agents,
    \item target-agnostic semantic navigation -- finding objects and object classes never seen during policy training,
    \item a collision-free navigation policy for complex, unseen environments that bridges the Sim2Real gap without any sort of retraining
\end{enumerate}

%% file: sections/2_approach.tex
\section{Approach}
\label{sec:approach}
\begin{table}
    \centering\vspace{1em}
    \caption{NavACL Geometric Properties}
    \label{tab:acl_feats}
    {\addtolength{\tabcolsep}{-5pt}\begin{tabular}{p{0.3\linewidth}p{0.68\linewidth}}\hline
         Geodesic Distance & The shortest-path distance from $s_0$ to $s_g$\\
         Path Complexity & The ratio of euclidean distance to geodesic distance of $s_0,s_g$\\
         Turn Angle & The angle between the starting orientation and $\overrightarrow{s_0s_g}$, represented as sine and cosine components\\
         Agent/Goal Clearance & Distance from $s_0$ and $s_g$ respectively to the nearest obstacle\\
         Agent/Goal Island & Distance from the centroid of the navigation mesh to its furthest connected edge at $s_0$ and $s_g$ respectively\\\hline
    \end{tabular}}
    \vspace{-1em}
\end{table}
\begin{figure*}[t]
    \centering\vspace{0.4cm}
    \includegraphics[width=0.72\linewidth]{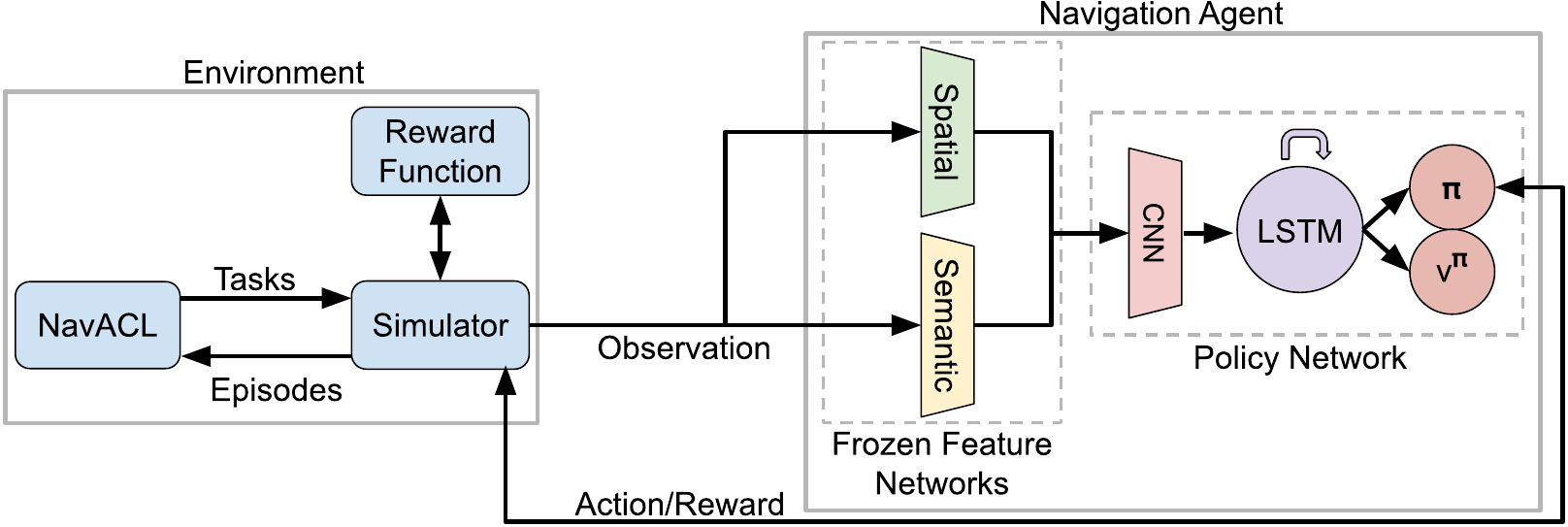}
    \caption{The agent training pipeline, and how our contributions fit within it. We compress observations from the environment into latent features before passing them to the policy network. NavACL trains on navigation episodes and serves tasks back to the simulator.}
    \label{fig:flowchart}\vspace{-1em}
\end{figure*}
Fig. \ref{fig:flowchart} presents a flowchart of our contributions, which includes NavACL, the reward function for collision-avoidance, and the frozen feature networks. We discuss each piece in the following subsections.
\subsection{NavACL}
NavACL is based on evidence that intermediate difficulty tasks provide more learning signal than random tasks for policy improvement \cite{portelas2020automatic,florensa2018automatic} and that replaying easy tasks alleviates catastrophic forgetting \cite{isele2018selective,rolnick2019experience,atherton2015memory}. NavACL filters down uniform random tasks to those that provide the most learning signal to the agent using predicted task success probability, described below.

Since our navigation problem has well-defined termination scenarios (agent reached the goal or not), we use binary task success as the signal metric. Let task $h = (s_0, s_g)$, with agent start position $s_0$ and goal position $s_g$. $f^*_\pi(h)$ denotes the probability of navigation policy $\pi$ solving task $h$, zero for certain failure and one for certain success. We estimate task success probability $f^*_\pi$ using a fully-connected deep neural network we call $f_\pi$. Before each forward pass, $f_\pi$ preprocesses $h$ into geometric properties (Tab.~\ref{tab:acl_feats}), allowing $f_\pi$ to generalize across scenes. We update $f_\pi$ alongside $\pi$ in the training loop (Alg. 1,2). We do not pretrain $f_\pi$ -- it learns only from rollouts produced by our agent, initially providing random tasks but quickly determining regions of varying task difficulty. We define task \emph{difficulty} as the complement of the estimated success probability, $1 - f_\pi(h)$. In contrast to \cite{florensa2018automatic} which formulates scenario generation with GANs for general frameworks, we optimize NavACL to generate scenarios efficiently for the navigation task using simple log loss.

\begin{figure}[t]
    \centering
    \includegraphics[width=\linewidth, trim=0 8em 0 8em, clip]{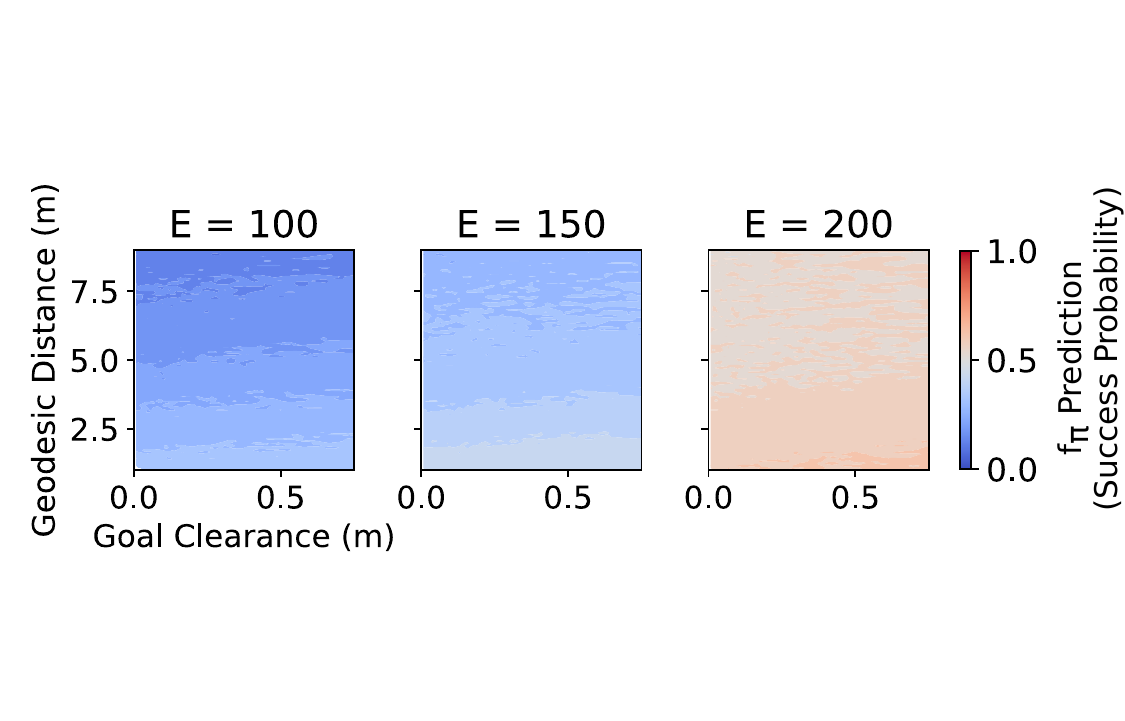}
    \caption{Spatial interpolation of $f_\pi$ across two geometric properties, at various training epochs E. Five thousand tasks are drawn randomly during each epoch and classified using $f_\pi$. Longer tasks (geodesic distance) are initially tough for the policy. As the policy improves, it can handle longer and harder tasks. Adaptive NavACL accounts for this, shifting its task distribution as the policy adapts.}
    \label{fig:acl_viz}
\end{figure}

\noindent\textbf{Adaptive filtering}\hspace{1em}
Now that we can estimate the difficulty of tasks, which tasks should we feed the agent? In one implementation, we produce \emph{goals of intermediate difficulty} (GOID) \cite{florensa2018automatic}, selecting tasks bounded between two success probabilities. GOID ensures we never select tasks that are too easy or too hard. However, GOID does not explicitly deal with catastrophic forgetting of easy tasks. Furthermore, the bounds do not change as the agent improves and task distribution shifts (Fig. \ref{fig:acl_viz}). Instead, we provide a mixture of task types, where certain tasks \emph{adapt} to the learner. \emph{Easy} tasks provide adequate learning signal early in training and prevent catastrophic forgetting. \emph{Frontier} tasks teach the agent to solve new tasks at its current ability. Uniformly sampled \emph{random} tasks inject entropy and prevent overfitting to specific task types. Initially, easy and frontier tasks form the majority of the task mixture. The mixture decays into random sampling as the learning agent learns to generalize.

\begin{algorithm}[t]
\footnotesize
    \caption{Training loop with $f_\pi$ update}\label{alg:update}
    \SetKwFunction{Union}{Union}\SetKwFunction{Train}{Train}
    \SetKwFunction{Union}{Union}\SetKwFunction{Rollouts}{Rollouts}
    \SetKwFunction{Union}{Union}\SetKwFunction{PPO}{PPO}
    \SetKwFunction{Union}{Union}\SetKwFunction{FitNormal}{FitNormal}
    \SetKwFunction{Union}{Union}\SetKwFunction{Init}{Init}
    \SetKwFunction{Union}{Union}\SetKwFunction{GetRandomTasks}{GetRandomTasks}
    \SetKwFunction{Union}{Union}\SetKwFunction{GetDynamicTask}{GetDynamicTask}
    \SetKwFunction{Union}{Union}\SetKwFunction{RunEpisode}{RunEpisode}
    \SetKwInOut{Input}{input}\SetKwInOut{Output}{output}
    \Input{$\emptyset$}
    \Output{$\pi$}
    $\pi, f_\pi, \mu_f, \sigma_f \leftarrow$ \Init{}\; 
    \For{i $\gets 0$ \KwTo numEpochs}{
    $tasks, successes, states, actions, rewards \leftarrow$ \Rollouts{$\pi$, $f_\pi, \mu_f, \sigma_f$}\;
    $\pi \leftarrow \PPO(\pi, states, actions, rewards)$\;
    $f_\pi \leftarrow$ \Train{$f_\pi, tasks, successes$}\;
    $randomTasks \leftarrow \GetRandomTasks{}$\;
    $\mu_{f}, \sigma_{f} \leftarrow$ \FitNormal{$f_\pi, randomTasks$}\;
    }
    \Return $\pi$\;
\end{algorithm}%
\begin{algorithm}[t]
\footnotesize
    \caption{Rollouts}
    \SetKwFunction{Union}{Union}\SetKwFunction{Rollouts}{Rollouts}
    \SetKwInOut{Input}{input}\SetKwInOut{Output}{output}
    \SetKwFunction{Union}{Union}\SetKwFunction{GetDynamicTask}{GetDynamicTask}
    \Input{$\pi, f_\pi; \mu_f; \sigma_f;$}
    \Output{$rollouts$}
        \For{i $\gets 0$ \KwTo batchSize} {
            $task \leftarrow$ \GetDynamicTask{$f_\pi, \mu_f, \sigma_f$}\;
            $rollouts[i] \leftarrow $\RunEpisode{$\pi, task$}\;
            }
        \Return $rollouts$\;
\end{algorithm}%
\begin{algorithm}[t]
\footnotesize
        \caption{GetDynamicTask}
        \SetKwFunction{Union}{Union}\SetKwFunction{GetTaskType}{GetTaskType}
        \SetKwFunction{Union}{Union}\SetKwFunction{RandomTask}{RandomTask}
        \SetKwInOut{Input}{input}\SetKwInOut{Output}{output}
        \Input{Training timestep $t$; $f_\pi$; $\mu_f; \sigma_f$; Hyperparameters $\beta, \gamma$}
        \Output{Task $h$}
        $taskType \leftarrow{}$ \GetTaskType{$t$}\;
        \While{true}{
            $h \leftarrow$ \RandomTask{} \;
            \Switch{$taskType$}{
                \Case{easy}{
                    \If{$f_\pi(h) > \mu_f + \beta \sigma_f$}{
                        \Return $h$\;
                    }
                }
                \Case{frontier}{
                    \If{$\mu_f - \gamma \sigma_f < f_\pi(h) < \mu_f + \gamma \sigma_f$}{
                        \Return $h$\;
                    }
                }
                \Case{random}{
                    \Return $h$\;
                }
            }
        }
\end{algorithm}
We draw many random tasks and estimate their difficulty using $f_\pi$, producing a difficulty estimate across the task space. We fit a normal distribution $\mu_f, \sigma_f$ to this distribution (Alg. \ref{alg:update}). $\mu_f, \sigma_f$ form an adaptive boundary in task space, partitioning it into easy and hard regions, predicated on policy $\pi$. In particular, task $h$ is considered an \emph{easy} task if $f_\pi(h) > \mu_f + \beta \sigma_f$ and a \emph{frontier} task if $\mu_f - \gamma \sigma_f < f_\pi(h) < \mu_f + \gamma \sigma_f$, where $\beta, \gamma$ are hyperparameters. In other words, task difficulty is relative to the current ability of the agent -- if we expect $\pi$ to do better on task $h$ than an average task, it is easy. If $h$ is near the difficulty of the average task, straddling the adaptive boundary, we call it a frontier task. Intuitively, this should provide a more conservative mixture of tasks than pure random sampling, promoting stable learning in difficult environments. The full algorithm is detailed in Alg. 1-3.

\subsection{Reward Shaping}
Our reward function provides negative rewards to discourage collision and intrinsic rewards to encourage exploration and movement. We define it as:
\begin{equation}
    r(s) = \mathds{1}_{succ} + \delta ( -\mathds{1}_{coll} + \mathds{1}_{expl}) + 0.01 d.
\end{equation}
The binary indicator $\mathds{1}_{succ}$ is true upon reaching the target, and false otherwise. $\mathds{1}_{coll}$ is true upon collision and false otherwise. The hyperparameter $0 < \delta < 1$ controls the agent's affinity for learning exploration and motor skills compared to target-seeking behavior. $\mathds{1}_{expl}$ is an intrinsic reward for exploration. We keep a buffer of past agent positions over an episode, and provide a reward if the current position of the agent is some distance from all previous positions. We find that without the intrinsic exploration term, the agent falls into a local maxima of spinning in place to avoid the negative reward from collisions, which is difficult to escape. $d$ is the distance traveled in the current step, expressing the prior that the agent should be trying to cover as large a search area as possible. We do not use a distance to target term, as others have shown binary success to be sufficient \cite{midLevelReps2018}.

\subsection{Frozen Feature Networks}
Traditional visual DRL agents use an autoencoder to transform input RGB images into a latent representation, where the autoencoder is trained end-to-end with the policy network \cite{ha2018world,7487173}. End-to-end training can overfit the policy network to simulation artifacts, and hurt real-world transfer \cite{gordon2019splitnet}. We use \emph{spatial} autoencoders pretrained on real images \cite{midLevelReps2018} and freeze their weights to prevent overfitting to simulation renders during training. High-polygon meshes scanned by \cite{xiazamirhe2018gibsonenv} and photorealistic renderers provided by \cite{savva2019habitat} produce detailed enough visualizations to work with encoders trained on real-world datasets. Freezing also speeds up policy convergence, as the gradient backpropagates through fewer layers. 

\subsection{Semantic Target Network}
We produce the semantic target feature using a Mask R-CNN with an FPN backbone trained on the COCO dataset \cite{wu2019detectron2,lin2014microsoft}. We introduce a small postprocessing layer that enables swapping target classes without retraining. Given an image, the Mask R-CNN predicts a binary mask $M$ for each object class, along with its confidence. We extract the mask with target label $l$ and do scalar multiplication of the binary mask with the prediction confidence to get output $O$. 
\begin{equation}
    O(x,y) = P(M(x,y)_{label=l}).
\end{equation}
We can change $l$ at runtime to search for different target classes. Pixels of $O$ still contain shape information on the target object (\eg a ball will produce a round mask but a box will produce a square one). To prevent the downstream policy from overfitting to one specific object shape, as well as reduce latent size, we apply a max-pool operation to $O$ which is then stacked along with the other features into a latent representation, which is fed to the policy network.

%% file: sections/3_model.tex
\section{Model description}
Our learner model consists of an actor-critic model with policy $\pi(s)$ and value function $V^\pi(s)$ optimized using proximal policy optimization (PPO) with clipping (Tab.~\ref{tab:ppo_params}) \cite{schulman2017proximal,stable-baselines}. The policy network and value function take latent representations from the feature encoders as input, and produce an action and value estimate as output. The policy networks consist of feature-compression and memory sections. The feature-compression section compresses spatially-coherent latent features into a more compact representation using convolutional layers. Receiving features instead of full RGB images reduces time to train and the likelihood of overfitting to the simulator. 

To keep the navigation problem Markovian, the state must contain information on where the agent has been, and if it has previously seen the target. The purpose of the memory section is to store this information. The memory section uses long short-term memory (LSTM) \cite{hochreiter1997long} cells to represent state in the partially observable Markov decision process (POMDP) \cite{sutton2018reinforcement}. With this, we aim to reduce the likelihood of revisiting previously explored areas and to remember the target location if it leaves the view. While rotating to circumvent obstacles, the agent may lose sight of the target.

\begin{table}
\setlength{\tabcolsep}{3pt}
    \centering
    \vspace{1em}
    \caption{PPO Parameters}
    \label{tab:ppo_params}
    \begin{tabular}{lrclr}
        \cline{1-2}\cline{4-5}
        \# of Minibatches & 1 &~& Learning Rate & 0.005\\
        Clipping Range $(\epsilon)$ & 0.10 && Discount Factor $(\gamma)$ & 0.99 \\
        Value Function Coef. $(c_1)$ & 0.5 && Entropy Coef. $(\beta \text{ or } c_2)$ & 0.01 \\
        Timesteps per Update & 4000 && Rollout Workers & 12\\
        Inner-Loop Epochs & 4 && GAE $\lambda$ & 0.95\\
        \cline{1-2}\cline{4-5}
    \end{tabular}
\end{table}

%% file: sections/4_experiments.tex
\section{Experiments}
\label{sec:experiments}

\begin{figure*}[t]
    \centering
    \begin{subfigure}{0.3\linewidth}
        \centering
        \includegraphics[width=\linewidth]{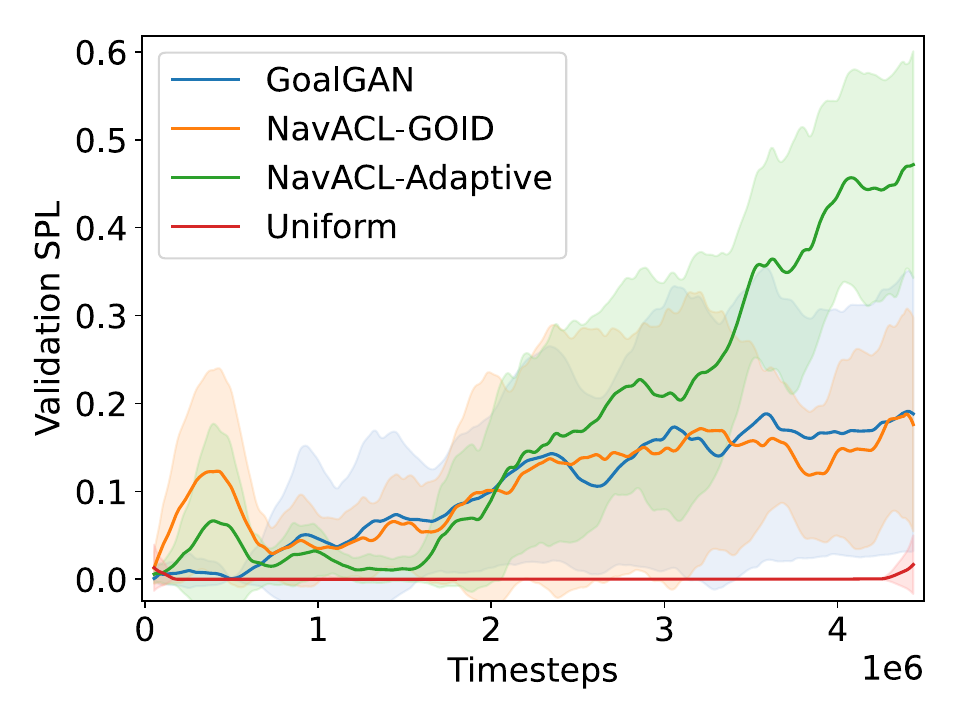}  
        \caption{Validation SPL}
        \label{fig:acl_ablation}
    \end{subfigure}
    \begin{subfigure}{0.3\linewidth}
        \centering
        \includegraphics[width=\linewidth]{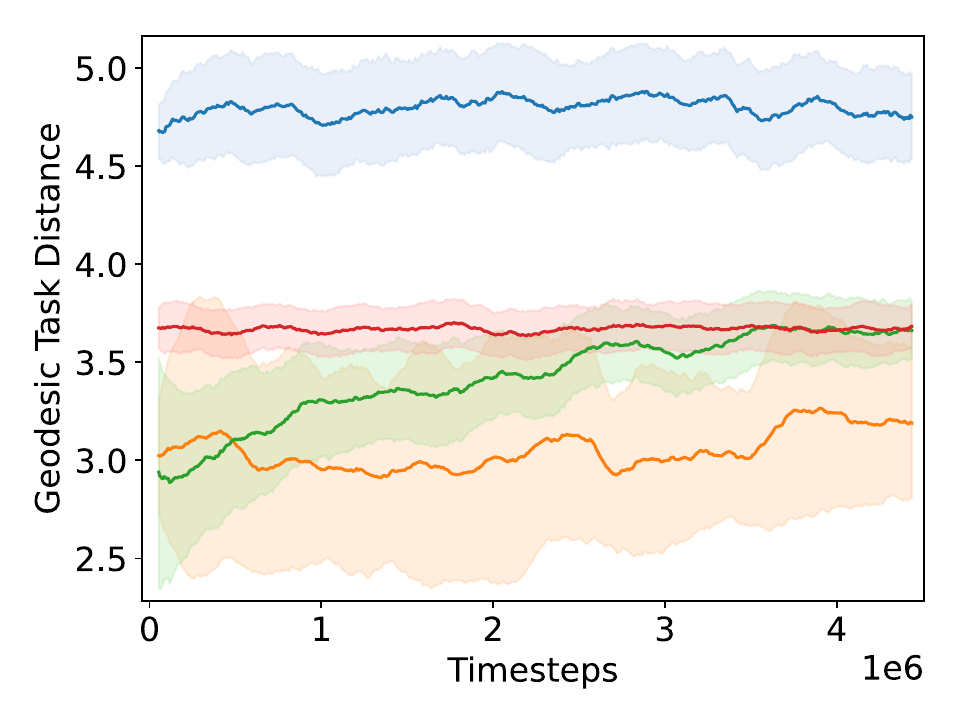}
        \caption{Task Selection}
        \label{fig:acl_tasks}
    \end{subfigure}
    \begin{subfigure}{0.3\linewidth}
        \centering
        \includegraphics[width=\linewidth]{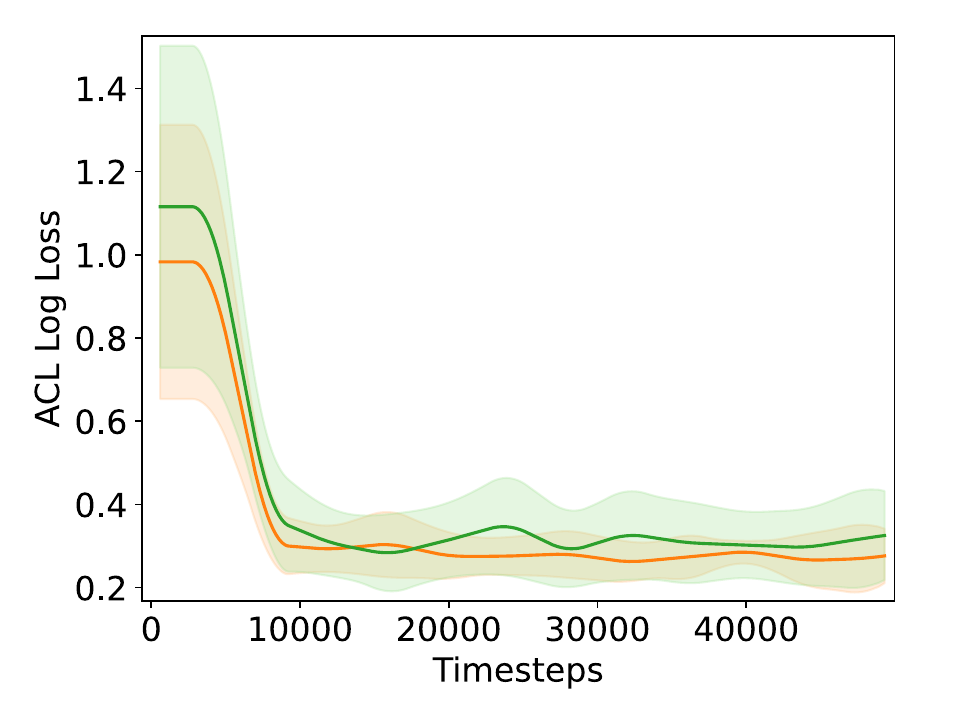}
        \caption{NavACL Convergence}
        \label{fig:acl_loss}
    \end{subfigure}
    \caption{(\ref{fig:acl_ablation}) Validation SPL over five trials on a single test-train environment. Solid lines and shaded areas refer to mean and one standard deviation, respectively. (\ref{fig:acl_tasks}) As the policy improves over time, NavACL increases the distance from start to goal -- ratcheting up the task difficulty. (\ref{fig:acl_loss}) NavACL's simple architecture provides valid predictions very quickly.}
    \label{fig:acl_results}
\end{figure*}
\begin{figure}[t]
    \centering
        \includegraphics[angle=90,height=0.21\linewidth]{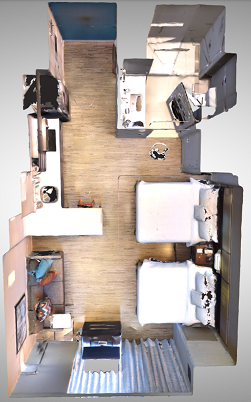}
        \hspace{0.01\linewidth}
        \includegraphics[height=0.21\linewidth]{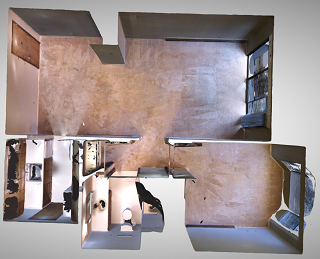}
        \hspace{0.01\linewidth}
        \includegraphics[height=0.21\linewidth]{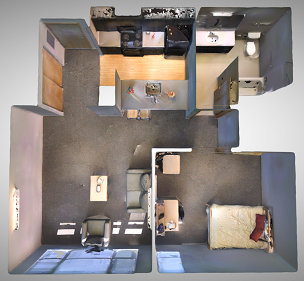}\hfill
    \caption{Simulation test scenes, left to right: Cooperstown (40m$^2$), Avonia (60m$^2$), Hometown (61m$^2$)}
    \label{fig:envs}
\end{figure}

We present three experiments: an ablation study of NavACL, a simulation benchmark of our model on unseen environments and target objects, and a benchmark of our agent operating in the real world.
\subsection{Evaluating NavACL}
Our first experiment compares the impact of NavACL on visual navigation to GoalGAN as well as the current standard of uniform random task sampling. We evaluate NavACL with GOID (NavACL-GOID) and with adaptive filtering (NavACL-Adaptive). We hold all policy parameters the same, and run five navigation trials of five million samples on the Cooperstown environment from the Gibson dataset \cite{xiazamirhe2018gibsonenv}. Uniform sampling uses Habitat's built-in task generator to generate tasks \cite{savva2019habitat}. GoalGAN uses an intermediate difficulty value between $0.1$ and $0.9$, used in their MazeAnt navigation experiment. When GoalGAN selects points outside of the scene, we select the closest valid point to it within the scene. We do not pretrain GoalGAN on past experiments -- we instead use the random initialization provided by the GoalGAN library, similar to NavACL initialization. For NavACL-GOID, we filter uniformly random tasks using our $f_\pi$ framework, and target tasks with an intermediate difficulty value of $0.4 \leq f_\pi(h) \leq 0.6$. NavACL-Adaptive uses hyperparameters $\beta = 1, \gamma = 0.1$. For NavACL-Adaptive, \texttt{GetTaskType} uses random task probability $\min(0.15,t/t_{max})$, where $t_{max}$ is the maximum number of training timesteps. Frontier and easy tasks split the remaining probability equally. NavACL-Adaptive samples 100 tasks to estimate $\mu_f$ and $\sigma_f$. In our reward function, we use $\delta = 0.25, d = 0.2$. 

NavACL-GOID performs on-par with GoalGAN, both outperforming uniform sampling. NavACL-Adaptive performs best, beating both NavACL-GOID and GoalGAN by a sizeable margin (Fig. \ref{fig:acl_results}). GoalGAN produces difficult tasks with long paths (Fig. \ref{fig:acl_tasks}), suggesting it has not learned the spatial layout of the complex scene (\eg where obstacles lie). Its performance being on-par with NavACL-GOID suggests that the adaptive portion of NavACL-Adaptive is responsible for the performance disparity.

\begin{figure*}
    \centering\vspace{0.3cm}
    \begin{subfigure}{0.27\linewidth}
        \centering
        \includegraphics[height=3.8cm]{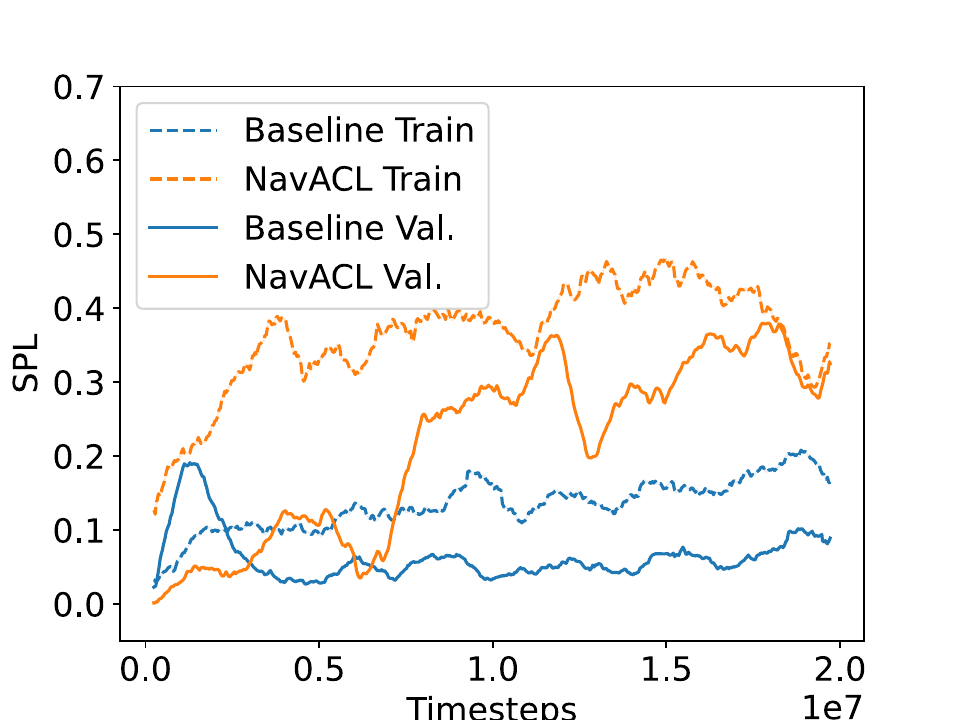} 
        \caption{SPL}
        \label{fig:train_results}
    \end{subfigure}
    \begin{subfigure}{0.4\linewidth}
        \centering
        \includegraphics[height=3.8cm]{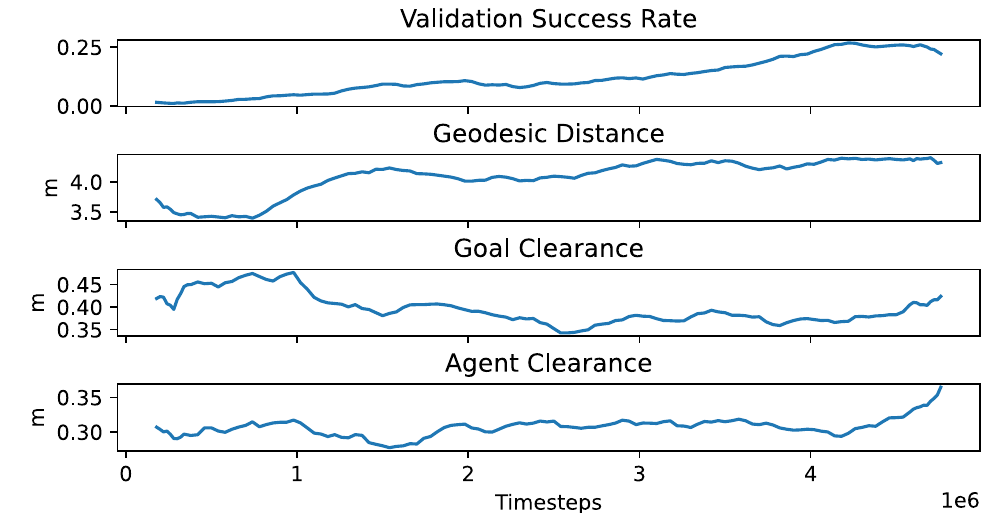}
        \caption{Geometric Feature Development}
        \label{fig:acl_multi}
    \end{subfigure}
    \begin{subfigure}{0.3\linewidth}
        \centering
        \includegraphics[height=3.8cm]{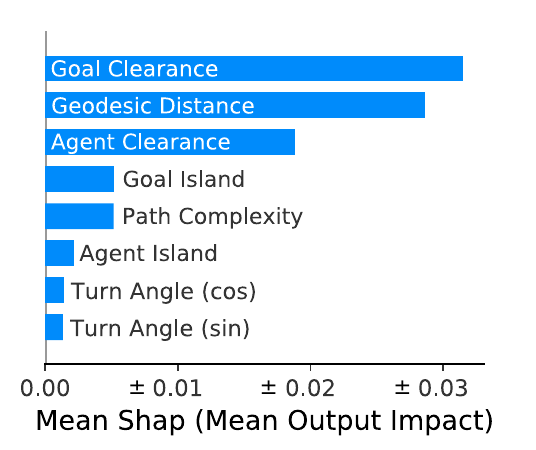}
        \caption{Feature Impact}
        \label{fig:acl_feat_ablation}
    \end{subfigure}
    \caption{(\ref{fig:train_results}) Training and validation results of our model trained with and without NavACL. (\ref{fig:acl_multi}) As the policy improves, NavACL produces harder tasks. Qualitatively, the agent learns obstacle avoidance motor skills before learning exploration behavior. Initially, NavACL selects shorter paths (short geodesic distance) in narrow corridors (low agent clearance) to guide the agent toward the goal. As the policy learns exploration strategies, the agent navigates without corridor guidance (high agent clearance), and even reaches goals behind obstacles (low goal clearance). (\ref{fig:acl_feat_ablation}) The mean prediction impact each feature had on NavACL's task prediction. Mean Shapley values \cite{shap} were determined using 1800 background samples, and feature impact was meaned over 200 test samples at 5M timesteps. At 5M timesteps, the agent has learned basic motion, so placing the target behind the agent (turn angle) has little impact. Path complexity has a smaller than expected impact, suggesting the measure could be improved.}
    \label{fig:results}
\end{figure*}

\subsection{Evaluating Model Performance}
Using our methodology, we train a policy over twenty million timesteps using the Habitat 2019 challenge split of the Gibson dataset. Policies are evaluated over ten trials of 30 episodes spread across three unseen test environments, held out from the Habitat split (Fig.~\ref{fig:envs}). The test tasks are randomly generated using the same uniform sampling as the Habitat challenge datasets \cite{savva2019habitat}. The target object is an 11cm radius football (soccer ball). All policies are limited to 150 moves, and all tasks have a maximum start to goal distance of 10m. We find increasing the number of moves beyond 150 results in little improvement. The action space consists of a rotation of $\pm 30^\circ$ and forward translation of $0.2$m.
\begin{table}[t]
    \centering
    \caption{Simulation results of collision-free agents over ten trials on unseen environments}
    \label{tab:sim_results}
    \begin{tabular}[t]{cccccc}
        \hline
        \multirow{2}{*}{Policy} & \multirow{2}{*}{Scene} & \multicolumn{2}{c}{Success Rate} & \multicolumn{2}{c}{SPL}\\
        & & $\mu$ & $\sigma$ & $\mu$ & $\sigma$\\
        \hline
        \multirow{4}{*}{Random} 
        & Cooperstown &  0.07 & 0.00 & 0.06 & 0.00\\
        & Avonia &  0.01 & 0.00 & 0.01 & 0.00 \\
        & Hometown & 0.00 & 0.00 & 0.00 & 0.00 \\
        & \textbf{Total} & \textbf{0.03} & \textbf{0.03} & \textbf{0.02} & \textbf{0.03} \\
        \hline
        \multirow{4}{*}{PPO Baseline} & Cooperstown & 0.41 & 0.12 & 0.16 & 0.05\\
        & Avonia & 0.28 & 0.09 & 0.12 & 0.04\\
        & Hometown & 0.23 & 0.10 & 0.13 & 0.05\\ 
        & \textbf{Total} & \textbf{0.31} & \textbf{0.13} & \textbf{0.14} & \textbf{0.05}\\
        \hline
        \multirow{4}{*}{\shortstack{NavACL \\Target-Agnostic}} & Cooperstown & 0.55 & 0.11 & 0.25 & 0.08\\
        & Avonia & 0.43 & 0.04 & 0.20 & 0.02\\
        & Hometown & 0.09 & 0.06 & 0.06 & 0.03\\
        & \textbf{Total} & \textbf{0.36} & \textbf{0.21} & \textbf{0.17} & \textbf{0.10}\\
        \hline
        \multirow{4}{*}{NavACL} & Cooperstown & 0.63 & 0.14 & 0.31 & 0.09\\
        & Avonia & 0.50 & 0.06 & 0.21 & 0.04\\
        & Hometown & 0.12 & 0.08 & 0.09 & 0.06\\ 
        & \textbf{Total} & \textbf{0.42} & \textbf{0.19} & \textbf{0.24} & \textbf{0.09} \\
        \hline
        \multirow{4}{*}{Human} & Cooperstown & 0.76 & 0.22 & 0.54 & 0.17\\
        & Avonia & 0.89 & 0.13 & 0.63 & 0.13\\
        & Hometown & 0.72 & 0.19 & 0.51 & 0.13\\ 
        & \textbf{Total} & \textbf{0.79} & \textbf{0.20} & \textbf{0.56} & \textbf{0.15} \\\hline
        \end{tabular}
\end{table}

The \textbf{Random} policy selects random actions to provide a lower bound on performance. The \textbf{PPO Baseline} policy is trained using depth, reshading (de-texturing and re-lighting), and semantic features, along with intrinsic rewards. The \textbf{NavACL} policy is trained identically to the PPO Baseline policy, with the addition of NavACL-Adaptive. \textbf{NavACL Target-Agnostic} is identical to NavACL, but during testing, we change the target from the ball to a large vase to evaluate target-agnostic semantic generalization to unseen targets of different shapes and sizes. We recruit ten volunteers from varying backgrounds to establish an upper-bound on performance. The \textbf{Human} policies are trained and tested just like the agent policies. The volunteers played the training set until they were comfortable with the controls, receiving the same RGB observations and action space as the agents. Once comfortable, the volunteers played through the same test set as the agents. We use the SPL metric defined by \cite{anderson2018evaluation} 
\begin{equation}
    \frac{1}{N} \sum_{i=1}^N \mathds{1}_{succ,i}  \frac{l_i}{\max(l_i, p_i)}
    \label{eqn:spl}
\end{equation}
where $l$ is the length of the agent's path, $p$ is the length of the shortest path from start to goal, and $N$ is the number of episodes. Results are presented in Fig. \ref{fig:results} and Tab. \ref{tab:sim_results}.

Our agents are able to find semantically specified targets in simulation, performing drastically better than random. Our NavACL methods improve upon PPO Baseline on average, and particularly in peak performance results of Cooperstown and Avonia. On Hometown, the NavACL agent is confused by a photoscanned mirror at the end of a long corridor, producing a mirage nearly 60 timesteps in length. The baseline policy behaves more randomly and falls for this trick less often. On unseen semantic classes, performance decreases slightly, but the policy shows target-agnostic semantic generalization capability. On average, humans outperform agents in unseen environments. However, humans can memorize test scenes during the first few episodes, giving them an advantage over agents during later episodes. On Cooperstown, agents are within one standard deviation of human-level performance in success rate, suggesting they outperform some humans in some cases. We found agents had trouble navigating to new spaces in larger, unseen environments. Agents did not have as much trouble when navigating in large, previously-seen environments. Memory for embodied visual agents is an active area of research \cite{gupta2017cognitive,gupta2017unifying,savinov2018semi,mezghani2020learning}, and we expect leveraging these memory modules will improve performance in larger environments. Another limitation was model throughput -- it took roughly two weeks to train twenty million timesteps on a GPU machine, and previous experiments were still showing policy improvement at sixty million timesteps. With memory improvements and distributed computing, we believe our models could approach human performance. 

\begin{figure*}
    \centering
    \begin{subfigure}{\linewidth}
        \centering\vspace{0.3cm}
        \includegraphics[width=0.7\linewidth,trim={0 0.38cm 0 0.0cm},clip]{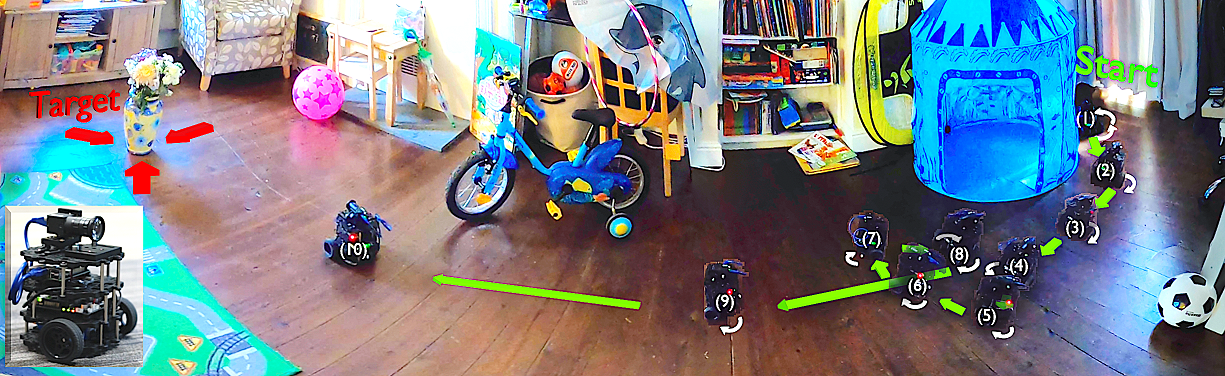}
        \caption{}
        \label{fig:pano_exp}
    \end{subfigure}
    \begin{subfigure}{\linewidth}
        \centering
        \includegraphics[width=0.7\linewidth]{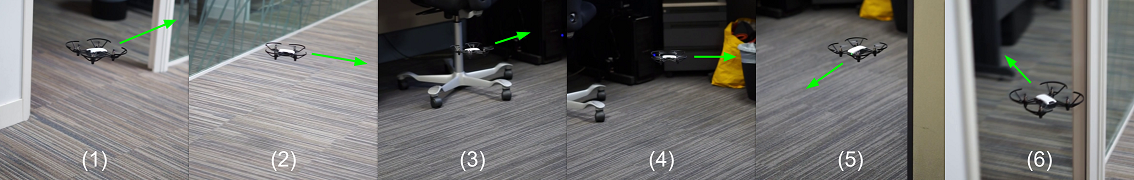}
        \caption{}
        \label{fig:droberto_action}
    \end{subfigure}
\caption{(\ref{fig:pano_exp}) The AGV navigating to the vase target in the house scene, with several never-before-seen obstacles littering the path to the target. Previous semantic targets (football and pink star ball) are present to emphasize target-agnostic semantic navigation capability. The agent turned 360$^\circ$ (1) to evaluate its options, then took the path between the desk and tent (2), adjusted the trajectory towards the wide-open area in front of the blue tent (3), and rotated 360$^\circ$ (4). The agent explored the areas surrounding the bike, bookshelves, and the blue tent (5,6,7). Target detection occurred at (8), and the AGV made a beeline for the target (9,10). (\ref{fig:droberto_action}) While flying down a hallway (1), the UAV notices an empty cubicle (2). It threads the needle, flying between the chair wheels and seat (3). After exploring the cubicle (4,5), it leaves and heads into the adjacent open office without collision (6). }
\label{fig:pano}
\end{figure*}

\subsection{Sim2Real Transfer}
\begin{table}[t]
    \centering
    \caption{Real-world results from the AGV}
    \label{tab:real_results}
    \begin{tabular}[t]{cccc}
    \hline
        Scene & Target & Task Dist. (m) & SPL\\
        \hline
        Office 1 & Football & 2.95 & 0.92 \\
        & Football & 3.09 & 0.53\\
        & Football & 2.92 & 0.42\\
        & Football & 1.67 & 0.18\\
        \hline
        Office 2 & Football & 9.10  & 0.65\\
        \hline
        House & Orange Football & 4.19 & 0.92\\
        \hline
        House & Vase & 5.14 & 0.54\\
        \hline
    \end{tabular}
    \vspace{-1.5em}
\end{table}
We transfer our policy without modification to a Turtlebot3 wheeled robot (AGV) and a DJI Tello quadrotor (UAV). The AGV uses wheel encoders for closed-loop control for motion primitives (single actions), but does not estimate odometry across actions. We tested the AGV on seven tasks spanning three environments and three objects, one being an unseen object and one being from an unseen semantic class. We use wheel odometry to measure SPL for the AGV (Tab. \ref{tab:real_results}). The AGV did not experience a single collision over the 29m it traveled during tests and was robust to actuator noise as well as wheel slip caused by terrain (hardwood, carpet, and rugs). 

The UAV uses IR sensors to determine its height and an IMU to obtain very noisy position estimates for motion primitives and hovering stability. \emph{We did not train a separate model for the UAV}. We used the model trained with the AGV height and camera field of view, which drastically differ from the UAV ($0.2$m vs $\sim 0.75-1.5$m and $68^\circ$ vs $47^\circ$ respectively). Policies trained for the AGV seemed surprisingly effective on the UAV, suggesting greater model generalization than we anticipated. The UAV was able to fly in-between legs of a camera tripod, through doorways, and even around moving people on many occasions without a single collision. Unfortunately, hover instability resulted in varying target height making target navigation unstable. While small changes in height were tolerated, larger differences were regarded as spurious detections. Still, this surprising generalization implies it may be possible to train a single navigation model for use on diverse types of robots that implement similar motion primitives. We provide video results of both the AGV and the UAV in the supplementary material,\footnote{Also available at \url{https://www.youtube.com/playlist?list=PLkG_dDkoI9pjPdOGyTec-sSu20pB7iayC}} and illustrations in Fig. \ref{fig:pano}.

%% file: sections/5_conclusions.tex
\section{Conclusions}
\label{sec:conclusions}

We introduce NavACL and present two variants, NavACL-GOID and NavACL-Adaptive, for navigation task generation. Both methods significantly improve upon uniform sampling (the current standard) as well as GoalGAN in sparse-reward settings. Combining NavACL with frozen feature networks and collision-free policies produces agents capable of target-agnostic semantic navigation in simulation and the real world.

\subsection{Future Work}
We found LSTMs had issues with generalization to new environments with the compute power available to us. Future work will focus on integrating more structured and efficient memory modules \cite{gupta2017cognitive,gupta2017unifying,savinov2018semi,mezghani2020learning} into our learning pipeline.

The unexpected real-world generalization ability between mobility types warrants further investigation. Training an agent with an actuator abstraction layer allows transfer to disparate, never-before-seen robots. It may be prudent to invest computational resources in training a single model with abstract actuation that can be applied to drones, wheeled robots, walking robots, blimps, etc., rather than train each model individually.

We evaluated NavACL using on-policy reinforcement learning, but NavACL may be useful for selecting which episodes to replay when using off-policy methods. It may also prove useful in selecting and ordering training episodes for imitation learning.